\documentclass[conference,a4paper]{IEEEtran}
\IEEEoverridecommandlockouts

\usepackage[hidelinks]{hyperref}
\usepackage[cmex10]{amsmath}%American Math Society(AMS) math formatting
\usepackage{amssymb,amsfonts}%AMS extra symbols and fonts
\usepackage{dblfloatfix}%fix double column figure ordering and placement

\usepackage[ruled,vlined]{algorithm2e}
\usepackage{graphicx}
\graphicspath{{Figures/PDF/}{Figures/PNG/}}

\usepackage{booktabs}
\usepackage{multirow}
\usepackage{siunitx}
\usepackage[numbers,compress]{natbib}
\usepackage{texnames}
\usepackage{bm,bbm}
\usepackage{orcidlink}

\begin{document}

\title{\uppercase{SharpSplat: Edge-Regularized 3D Gaussian Splatting for High Fidelity Urban Building Reconstruction from UAV images}}

\author{	\IEEEauthorblockN{Porus Vaid\orcidlink{0000-0001-8632-7167}}
	\IEEEauthorblockA{\textit{IISER Bhopal}\\
		Madhya Pradesh, India\\
		porus21@iiserb.ac.in}
	\and
	\IEEEauthorblockN{Shivam Chopra\orcidlink{0009-0004-6126-1367}}
	\IEEEauthorblockA{\textit{IISER Bhopal}\\
		Madhya Pradesh, India\\
		shivamc21@iiserb.ac.in}
	\and
	\IEEEauthorblockN{Vaibhav Kumar*\orcidlink{0000-0002-0047-0681}}
	\IEEEauthorblockA{\textit{IISER Bhopal}\\
		Madhya Pradesh, India\\
		vaibhav@iiserb.ac.in (corresponding author)}
}

\maketitle
\begin{abstract}
Reconstructing high-fidelity 3D building models from UAV imagery is essential for large-scale digital twin development. However, existing 3D Gaussian Splatting (3DGS) techniques often struggle with building facades, failing to capture sharp geometric transitions. To address this, we propose a semantic edge regularization framework that supervises 3DGS to produce crisp architectural boundaries. Our method leverages SAM 3 to generate precise building masks, from which we extract architecturally significant edges. During training, we align rendered image gradients with these extracted edges, forcing the Gaussians to converge into sharp structural geometries. Evaluations across campus environments, dense urban centers, and custom residential datasets demonstrate significant improvements in edge fidelity without requiring architectural modifications to the 3DGS pipeline. Our approach proves robust across diverse building types, roof geometries, and urban densities.
\end{abstract}


\section{Introduction}

Large-scale 3D reconstruction from UAV imagery has become essential for urban planning, infrastructure monitoring, and digital twin applications. The reconstruction task is even more challenging in complex cities in developing countries like India. Current methods span various approaches: multi-view stereo~\cite{schonberger2016sfm}, neural radiance fields~\cite{mildenhall2020nerf}, and point-based representations, but 3D Gaussian Splatting~\cite{kerbl20233dgs} has emerged as the leading technique. Its explicit primitive representation enables fast training and real-time rendering while achieving photorealistic quality. Urban reconstruction and digital twin creation particularly benefits from this efficiency when processing hundreds of UAV images covering vast area.

Recent large-scale 3D reconstruction methods CityGaussianV2~\cite{xu2024citygaussianv2}, VastGaussian~\cite{lin2024vastgaussian}, and GS4Buildings~\cite{zhang2025gs4buildings} handle massive urban scenes efficiently. To simulate autonomous systems and urban planning, generating building twins with accurate facades has become necessary, as current Gaussian-splatting-based methods cannot reproduce accurate 3D geometries. This occurs because Gaussian primitives' smooth spatial transitions through their continuous falloff functions. Standard optimization minimizes global photometric error without explicit constraints at boundary geometry. The 3D models look acceptable from a distance but fail under close inspection, building facades blur into surroundings, window patterns lose definition, roof corners become rounded.

Recent work explores edge awareness in 3D reconstruction. Methods like DET-GS~\cite{detgs2025} and DN-Splatter~\cite{turkulainen2025dnsplatter} use depth maps to guide where Gaussians are placed, but depth estimation from UAV imagery often fails. Wide camera baselines cause correspondence problems, repetitive building textures confuse matching algorithms, and computational costs explode with scene size. EdgeGaussians~\cite{lei2024edgegaussians} introduces specialized primitives for boundaries, requiring substantial changes to the reconstruction pipeline. These approaches either break down on large scenes or need too much engineering for modest improvements in 3D building twin quality.

We take a different path. Rather than modifying the 3D representation or relying on unreliable depth, we supervise the 3D reconstruction process using semantic boundary information. From SAM3~\cite{sam3}, we generate binary building masks from UAV images using text prompts. We apply edge detection to the original images, then use these masks to isolate edges that belong to buildings roofs, facades, and corners. During 3D Gaussian Splatting optimization, we supervise the training such that when rendered from any viewpoint, these building edges appear sharp. This forces the Gaussians to arrange themselves in 3D space to produce crisp boundaries. The primitives learn positions and shapes that create well-defined building geometry rather than smooth, blurred transitions.

\noindent Our contributions are:
\begin{itemize}
    \item A method to accurately \textbf{extract building edges from UAV imagery} using SAM3 semantic segmentation combined with image edge detection, isolating architecturally meaningful boundaries.
    \item An edge regularization loss integrated into 3DGS optimization that supervises 3D Gaussian arrangement to produce sharp building edges in the reconstructed 3D model, \textbf{requiring no architectural modifications.}
    \item Validation across diverse scenes like residential buildings, commercial structures, university campuses, and dense urban areas with varied roof geometries and architectural styles, demonstrating our method's generalization capability.
\end{itemize}

\section{Related Work}

\begin{figure*}[t]
	\centering
	\includegraphics[width=\textwidth]{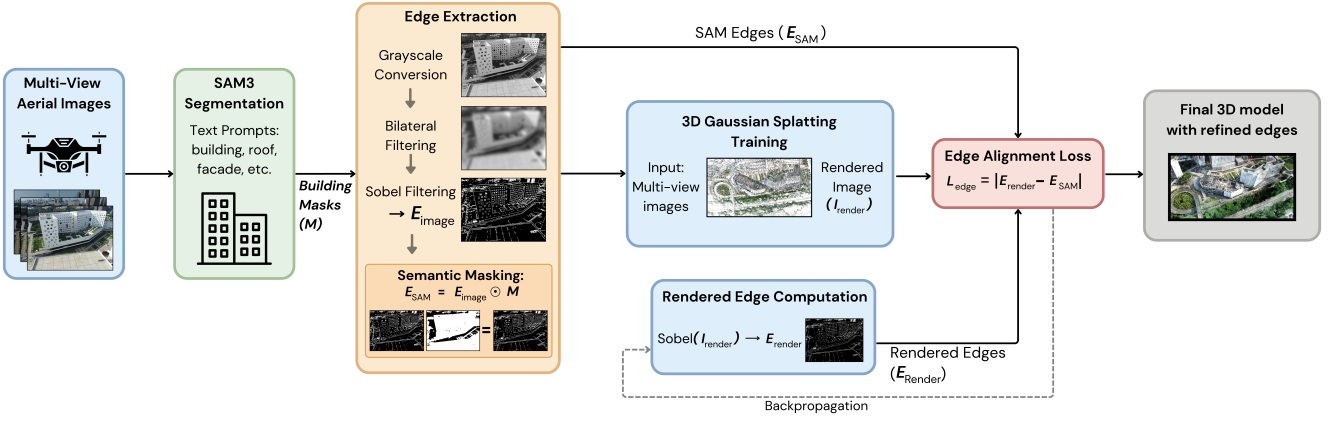}
	\caption{Overview of our semantic edge supervision approach. Multi-view images are processed through SAM3 to extract building masks using text prompts. Edge maps are generated by applying Sobel filtering to the original images and retaining only edges within masked building regions. During 3D Gaussian Splatting training, rendered images produce their own edge maps through identical Sobel filtering, which are aligned to SAM3 edges through our edge alignment loss $\mathcal{L}_{\text{edge}}$. This provides explicit supervision at semantic boundaries without architectural modifications.}
	\label{fig:method}
\end{figure*}

\subsection{3D Gaussian Splatting}

3D Gaussian Splatting~\cite{kerbl20233dgs} introduced an efficient alternative to NeRF~\cite{mildenhall2020nerf} for novel view synthesis. The method represents scenes as 3D Gaussian primitives with learnable parameters (position, covariance, opacity, color). Differentiable rasterization enables fast training and real-time rendering. Extensions have improved various aspects: Mip-Splatting~\cite{yu2024mipsplatting} that addresses aliasing, 2D Gaussian Splatting~\cite{huang20242dgs} using planar primitives for better surfaces reconstruction. For large-scale scenes, CityGaussian~\cite{liu2024citygaussian} and VastGaussian~\cite{lin2024vastgaussian} partition space to handle massive datasets. These methods focus on geometric accuracy, rendering speed, or scale. None specifically addresses edge sharpness in 3D appearance.

\subsection{Reconstructing digital twins of buildings}

Building reconstruction from UAV images presents unique challenges in preserving accurate building facades in digital twins. GS4Buildings~\cite{zhang2025gs4buildings} adapts 3DGS for building-centric reconstruction with geometric regularization. PG-SAG~\cite{pgsag2025} uses planar-based supervision on surface normals to improve wall flatness. SF-Recon~\cite{xiong2024sfrecon} tackles large-scale urban scenes with surface-aware primitives. These approaches constrain \textit{geometry}, ensuring walls are flat, roofs are planar. Our work is complementary: we constrain \textit{appearance} at edges regardless of underlying geometry.

\subsection{Edge-Aware Reconstruction}

Recent 3DGS extensions explore edge awareness in reconstruction. DET-GS~\cite{detgs2025} and DN-splatter~\cite{turkulainen2025dnsplatter} use depth and normal estimates to guide Gaussian densification for better 3D boundary geometry. However, depth estimation proves unreliable for UAV images due to wide baselines and texture ambiguities. EdgeGaussians~\cite{lei2024edgegaussians} introduces specialized edge primitives for 3D boundary representation, requiring substantial architectural changes to the splatting framework. These methods focus on where to place Gaussians in 3D space. We focus on how those Gaussians should be configured to produce sharp edges when the 3D model gets rendered from any viewpoint, directly supervising the reconstruction process rather than the primitive placement strategy.

\subsection{SAM in 3D Reconstruction}

The Segment Anything Model~\cite{kirillov2023sam} has found applications in 3D tasks. SAGA~\cite{cen2024saga} and Gaussian Grouping~\cite{ye2024gaussiangrouping} use SAM masks to learn 3D semantic features for object manipulation. SA-GS~\cite{cen2024sags} applies masks for spatial regularization, controlling how Gaussians distribute within segmented regions. These works use SAM for identifying objects in 3D scenes. We use SAM3~\cite{sam3} with text prompts generate binary building masks, we apply edge detection to training images and use masks to isolate building edges, then supervise 3D reconstruction so these edges appear sharp in the final model. SAM3 provides semantic context about what constitutes a building, while our supervision targets achieving sharp 3D geometry at those building boundaries.

\section{Method}

Our approach integrates semantic edge supervision into standard 3DGS training without modifying the underlying architecture. The method consists of three components: SAM-based edge extraction, rendered edge computation, and edge alignment loss.

\subsection{SAM-Based Edge Extraction}

We employ SAM3~\cite{sam3}, a concept-based segmentation model that extends the Segment Anything Model~\cite{kirillov2023sam} with text-guided capabilities. Unlike SAM's point/box prompts, SAM3 accepts natural language descriptions, enabling precise building detection through descriptive text prompts. For each training image, we apply SAM3 with building-specific text prompts (\textit{``building''}, \textit{``concrete structure''}, \textit{``modern architecture''}, \textit{``white building facade''}, \textit{``roof''}) to segment building regions comprehensively. The resulting binary masks $M \in \{0,1\}^{H \times W}$ delineate building boundaries with high precision. \\

To extract edge information, we apply Sobel filtering to original training images rather than binary masks, preserving fine-grained features. Sobel outputs continuous gradient magnitudes which is differentiable for backpropagation and runs efficiently as Conv2d operations. After bilateral filtering for noise reduction, Sobel computation follows:
\begin{equation}
	G_x = \begin{bmatrix} -1 & 0 & 1 \\ -2 & 0 & 2 \\ -1 & 0 & 1 \end{bmatrix} * I_{\text{smooth}}, \quad
	G_y = \begin{bmatrix} -1 & -2 & -1 \\ 0 & 0 & 0 \\ 1 & 2 & 1 \end{bmatrix} * I_{\text{smooth}}
\end{equation}

The edge magnitude is computed as $E_{\text{image}} = \sqrt{G_x^2 + G_y^2}$. We then apply \textbf{semantic filtering} using the SAM3 derived building mask:
\begin{equation}
	E_{\text{SAM}} = E_{\text{image}} \odot M
\end{equation}

where $\odot$ denotes element-wise multiplication. We apply slight morphological dilation (5×5 kernel) to $M$ before masking to ensure edge coverage near boundaries. This produces semantically-aware edges capturing architectural features. 

\subsection{Rendered Edge Computation}

During training, we render RGB images $I_{\text{render}}$ through differentiable gaussian rasterizer. We compute edge maps from rendered images using the same Sobel process to ensure consistency. After converting to grayscale, we apply Sobel filtering directly without bilateral smoothing, as rendered images are already smooth by construction:
\begin{equation}
	E_{\text{render}} = \sqrt{(G_x * I_{\text{render,gray}})^2 + (G_y * I_{\text{render,gray}})^2}
\end{equation}

We normalize both $E_{\text{render}}$ and $E_{\text{SAM}}$ to [0,1] range to ensure balanced comparison.

\subsection{Edge Alignment Loss}

We formulate edge supervision as an L1 distance:
\begin{equation}
	\mathcal{L}_{\text{edge}} = \frac{1}{HW} \sum_{i,j} |E_{\text{render}}(i,j) - E_{\text{SAM}}(i,j)|
\end{equation}

This loss integrates into standard 3DGS training through weighted combination:
\begin{equation}
	\mathcal{L}_{\text{total}} = \mathcal{L}_{\text{RGB}} + \lambda_{\text{SSIM}} \mathcal{L}_{\text{SSIM}} + \lambda_{\text{edge}} \mathcal{L}_{\text{edge}}
\end{equation}

Edge loss computation executes efficiently through Conv2d operations with fixed 3×3 Sobel kernels. SAM3 edge maps are precomputed offline and loaded during training without gradient tracking, ensuring memory efficiency.

\begin{figure*}[t]
	\centering
	\includegraphics[width=\textwidth]{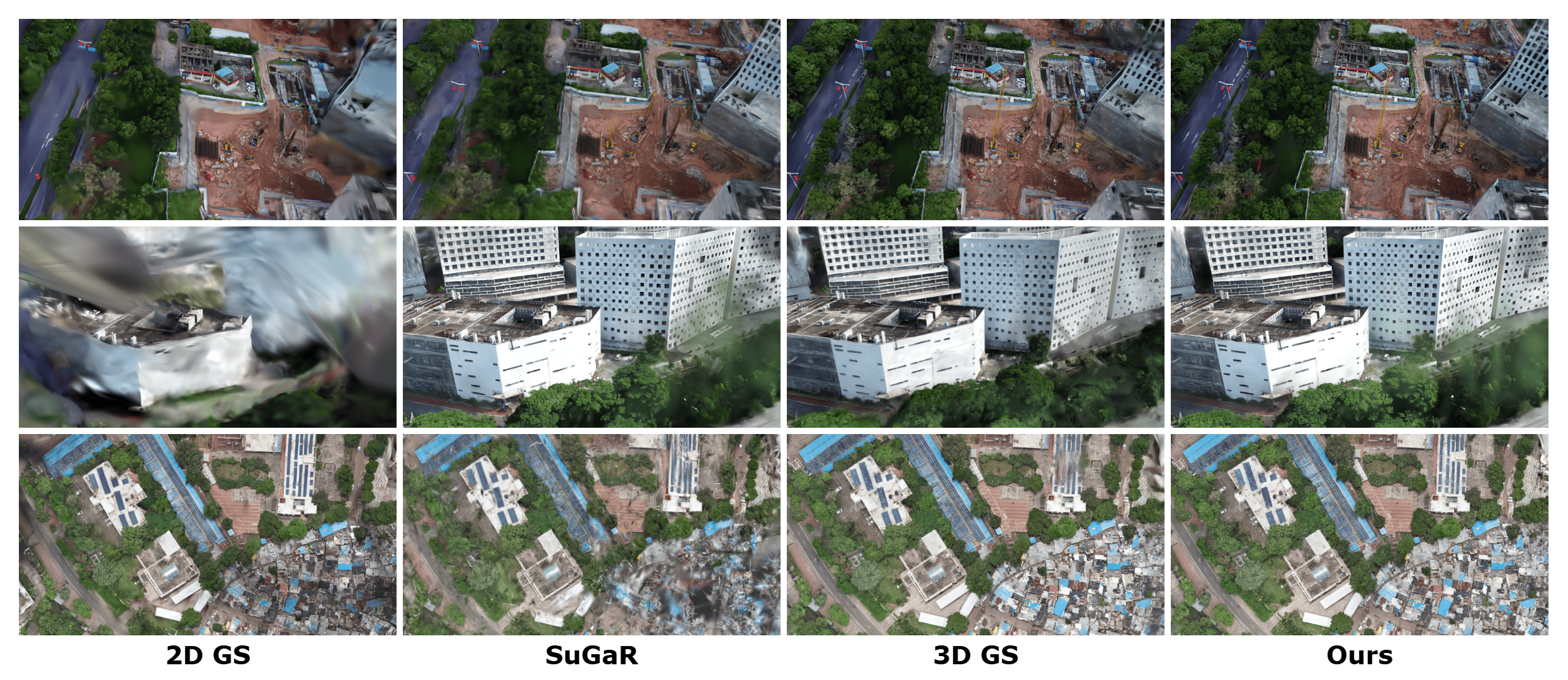}
	\caption{Comparison with SOTA methods across three scenes. Top row: Art Sci scene. Middle row: PolyTech scene. Bottom row: Gehukheda. From left to right: 2DGS, SuGaR, 3DGS baseline, and our edge-supervised method. Our approach produces sharper building edges and clearer architectural details compared to all baselines, including recent surface-based methods (2DGS, SuGaR) that improve geometry but not RGB edge appearance.}
	\label{fig:comparison_sota}
\end{figure*}

\begin{figure}[t]
	\centering
	\includegraphics[width=\columnwidth]{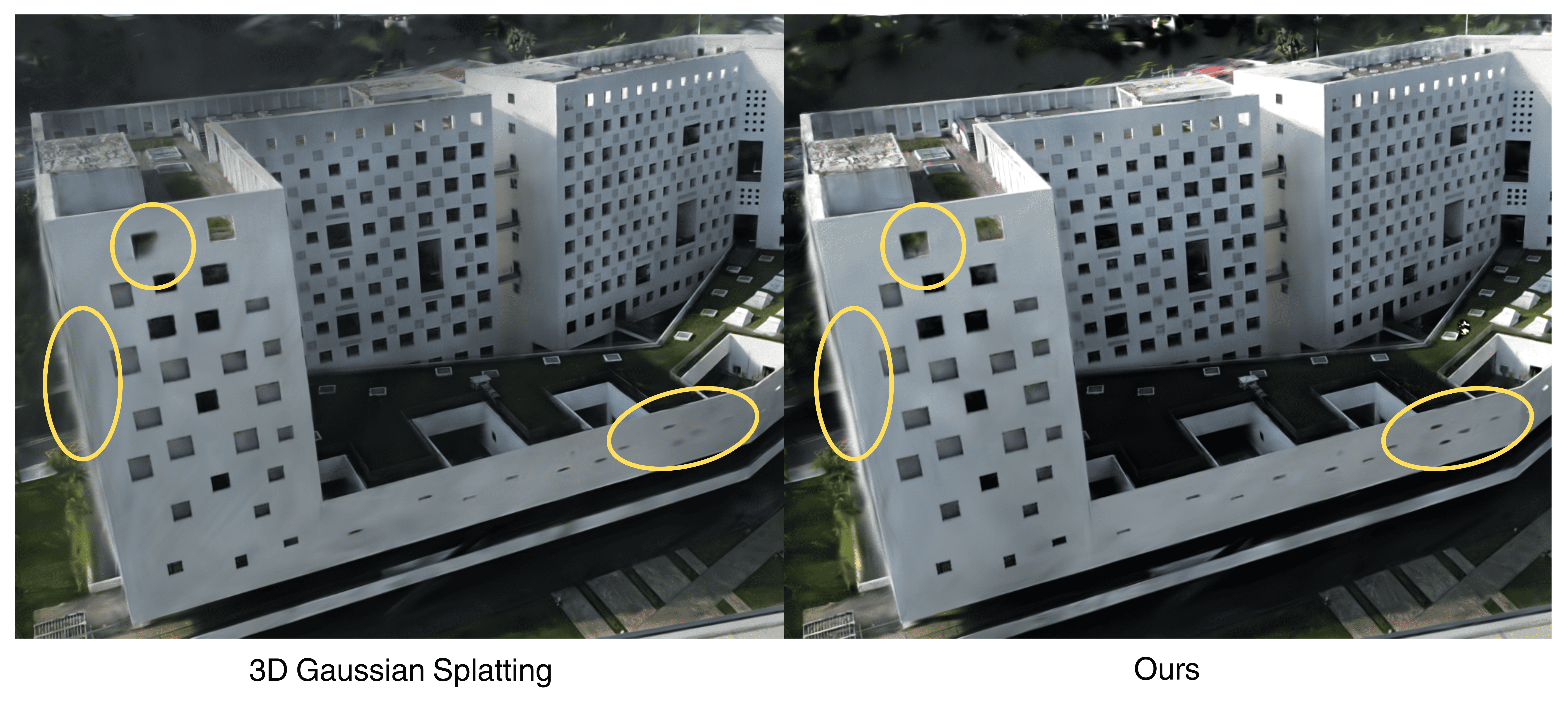}
	\caption{PolyTech scene comparison. Left: Baseline 3DGS with blurred edges (red circles). Right: Our method with sharp boundaries (blue circles). Architectural features including window patterns, building corners, and structural elements exhibit notably improved sharpness with edge supervision.}
	\label{fig:polytech1}
\end{figure}

\begin{figure}[t]
	\centering
	\includegraphics[width=\columnwidth]{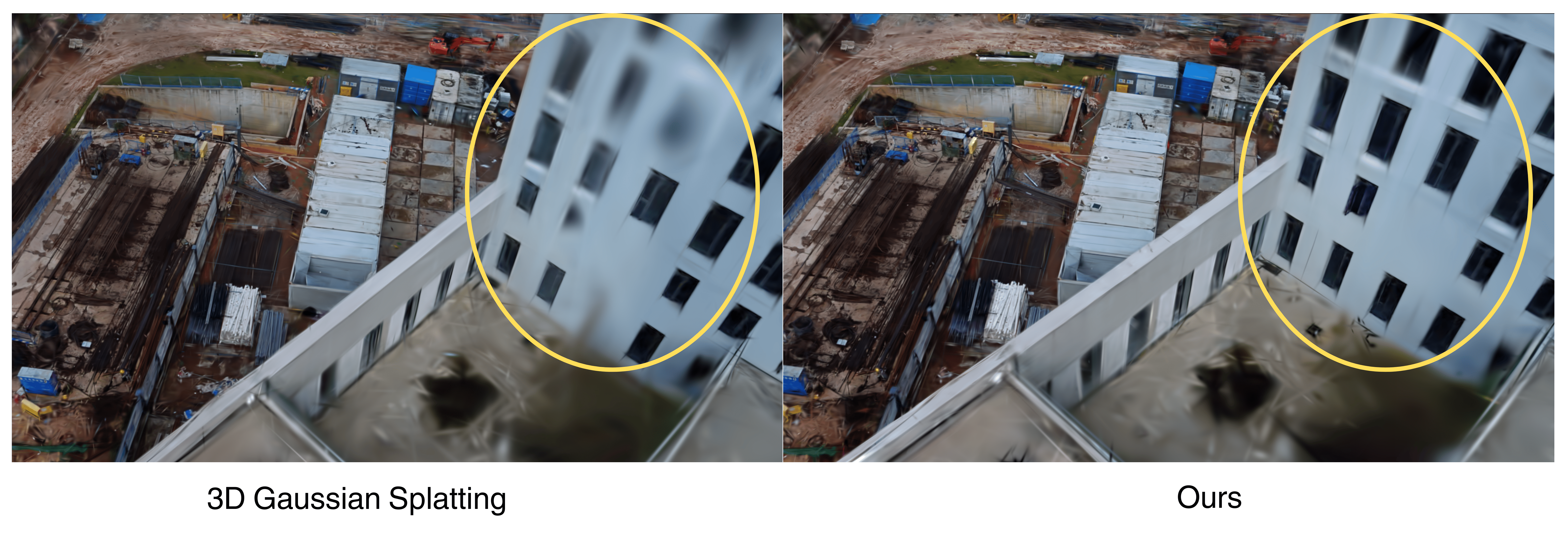}
	\caption{Art Sci scene comparison. Left: Baseline 3DGS. Right: Our method.  Edge supervision significantly improves window frame clarity.}
	\label{fig:artsci1}
\end{figure}

\section{Experiments and Results}

\subsection{Experimental Setup}

For training and evaluation we are using two UAV datasets, UrbanScene3D~\cite{urbanscene3d} which has scenes like PolyTech and Art Sci, and \textbf{Gehukheda} area of Bhopal which contains self-collected residential and commercial building imagery with diverse roof geometries. \textbf{PolyTech} (763 images) covers university campus buildings with varied architecture. \textbf{Art Sci} provides dense urban layouts with complex occlusions. All scenes processed at resolution factor 1/2.

We test at two training lengths: 7K and 15K iterations. This shows both convergence behavior and practical efficiency. Standard 3DGS parameters with adaptive density control, exponential learning rate decay, and $\lambda_{\text{SSIM}} = 0.2$. Camera poses from COLMAP~\cite{schonberger2016sfm}. SAM3 edge maps precomputed via text-prompted segmentation and Sobel filtering.

\begin{table}[hbt]
    \centering
    \caption{Results at 15K iterations across three datasets.}\label{tab:results_15k}
    \small
    \begin{tabular}{l l S[table-format=2.4] S[table-format=1.4] S[table-format=1.4]}
        \toprule
        \textbf{Dataset} & \textbf{Method} & \textbf{PSNR (dB)} $\uparrow$ & \textbf{SSIM} $\uparrow$ & \textbf{LPIPS} $\downarrow$ \\ 
        \midrule
        \multirow{2}{*}{PolyTech} & 3D GS & 19.4311 & 0.6565 & 0.4230 \\
        & Ours & \textbf{19.6277} & \textbf{0.6605} & \textbf{0.4160} \\
        \midrule
        \multirow{2}{*}{Art Sci} & 3D GS & 19.7642 & 0.4957 & 0.5861 \\
        & Ours & \textbf{19.8671} & \textbf{0.4995} & \textbf{0.5852} \\
        \midrule
        \multirow{2}{*}{Gehukheda} & 3D GS & 19.6096 & 0.5381 & 0.5389 \\
        & Ours & \textbf{19.8455} & \textbf{0.5389} & \textbf{0.5382} \\
        \bottomrule
    \end{tabular}
\end{table}

\begin{table}[hbt]
    \centering
    \caption{Results at 7K iterations across three datasets.}\label{tab:results_7k}
    \small
    \begin{tabular}{l l S[table-format=2.4] S[table-format=1.4] S[table-format=1.4]}
        \toprule
        \textbf{Dataset} & \textbf{Method} & \textbf{PSNR (dB)} $\uparrow$ & \textbf{SSIM} $\uparrow$ & \textbf{LPIPS} $\downarrow$ \\ 
        \midrule
        \multirow{2}{*}{PolyTech} & 3D GS & 17.5808 & 0.5576 & 0.5286 \\
        & Ours & \textbf{17.6803} & \textbf{0.5648} & \textbf{0.5175} \\
        \midrule
        \multirow{2}{*}{Art Sci} & 3D GS & 18.2584 & 0.4171 & 0.6847 \\
        & Ours & \textbf{18.3789} & \textbf{0.4186} & \textbf{0.6831} \\
        \midrule
        \multirow{2}{*}{Gehukheda} & 3D GS & 18.0760 & 0.4470 & 0.6643 \\
        & Ours & \textbf{18.1837} & \textbf{0.4483} & \textbf{0.6641} \\
        \bottomrule
    \end{tabular}
\end{table}

\subsection{Results}

Tables~\ref{tab:results_15k} and~\ref{tab:results_7k} show quantitative results. Edge supervision never degrades baseline performance improvements occur consistently across metrics and datasets. Memory requirements unchanged since edge maps load without gradients.

Comparison with 2DGS (a planar-primitive method) reveals complementary behavior. While 2DGS improves geometric accuracy through explicit surfaces, it doesn't address RGB edge appearance. Our method targets visual sharpness directly, suggesting these approaches could combine in future work. SuGaR similarly focuses on mesh extraction rather than edge rendering quality.

\section{Conclusion}

We introduced semantic edge supervision for 3DGS building reconstruction using SAM3 boundaries as guidance. The approach addresses a specific problem of edge blurriness in 3D reconstructed buildings. By supervising RGB gradients at semantically meaningful locations, we improve visual quality without architectural modifications to 3DGS.

Limitations of this method depends on SAM3 segmentation accuracy, and $\lambda$ requires per-scene tuning (though values cluster around 0.001-0.1).

The method adapts to different scene complexities through $\lambda$ tuning and maintains improvements even with reduced training time.

Our work is complementary to geometric methods like 2DGS and PG-SAG that constrain surfaces through depth or normals. We constrain appearance. Future work could combine both approaches, using geometric methods for structural accuracy and appearance based edge supervision for visual sharpness. The edge alignment framework could extend to other Gaussian-based reconstruction methods beyond 3DGS.

\small
\bibliographystyle{IEEEtranN}
\bibliography{references}

\end{document}